# Gated-Dilated Networks for Lung Nodule Classification in CT scans

**Mundher Al-Shabi[1], Hwee Kuan Lee[2,3,4], Maxine Tan[1,5]**
[1]Electrical and Computer Systems Engineering Discipline, School of Engineering, Monash University Malaysia, Bandar Sunway, Selangor, Malaysia
[2]Bioinformatics Institute, Agency for Science, Technology and Research (A*STAR), 30 Biopolis Street, 138671, Singapore
[3]Image and Pervasive Access Lab (IPAL), CNRS UMI 2955, 1 Fusionopolis Way, 138632, Singapore
[4]Singapore Eye Research Institute, 20 College Road, 169856, Singapore
[5]School of Electrical and Computer Engineering, University of Oklahoma, Norman, OK 73019, USA

Corresponding author: Mundher Al-Shabi (e-mail: mundher.al-shabi @monash.edu).

This work was supported by the Fundamental Research Grant Scheme (FRGS), Ministry of Education Malaysia (MOE), under grant FRGS/1/2018/ICT02/MUSM/03/1. This work was also supported by the Electrical and Computer Systems Engineering and Advanced Engineering Platform, School of Engineering, Monash University Malaysia. This work was also supported by the TWAS-COMSTECH Joint Research Grant, UNESCO.

**ABSTRACT** Different types of Convolutional Neural Networks (CNNs) have been applied to detect cancerous lung nodules from computed tomography (CT) scans. However, the size of a nodule is very diverse and can range anywhere between 3 and 30 millimeters. The high variation of nodule sizes makes classifying them a difficult and challenging task. In this study, we propose a novel CNN architecture called Gated-Dilated (GD) networks to classify nodules as malignant or benign. Unlike previous studies, the GD network uses multiple dilated convolutions instead of max-poolings to capture the scale variations. Moreover, the GD network has a Context-Aware sub-network that analyzes the input features and guides the features to a suitable dilated convolution. We evaluated the proposed network on more than 1,000 CT scans from the LIDC-LDRI dataset. Our proposed network outperforms state-of-the-art baseline models including Multi-Crop, Resnet, and Densenet, with an AUC of >0.95. Compared to the baseline models, the GD network improves the classification accuracies of mid-range sized nodules. Furthermore, we observe a relationship between the size of the nodule and the attention signal generated by the Context-Aware sub-network, which validates our new network architecture.

## I. INTRODUCTION

Lung cancer is the second most common cancer in men and women [1], [2]. In 2018 alone, there have been approximately 234,030 new cases and 154,050 deaths from lung cancer [2] altogether, which makes it by far the most common cause of cancer death among both men and women. Every year, more people die of lung cancer than of colon, breast, and prostate cancers combined [2]. Early diagnosis of lung cancer is extremely important for early treatment and cure of the disease. Early-stage lung cancer is typified by a small nodule, which can be detected as a round, spherical structure in computed tomography (CT) scans [3]. Doctors typically extract multiple features/characteristics of nodules from CT scans, such as size, morphology, contours, interval growth between CT examinations, multiplicity, location, and calcifications [4], [5]. These characteristics help to classify a nodule as benign (non-cancerous) or malignant (cancerous), e.g., malignant nodules frequently have more irregular boundaries/margins as compared to benign nodules, which generally have more smooth boundaries [5].

In recent years, Convolution Neural Networks (CNNs) have been gaining widespread popularity in general applications [6], [7], and have also been applied in recent studies [8]–[10] to classify/detect lung nodules. However, one of the main issues facing Computer-Aided Diagnosis (CAD) schemes for lung nodule detection and classification is the wide variation of nodule sizes. In our preliminary analysis, we observed that most nodules that are malignant tend to have a bigger nodule size/diameter than benign nodules [5]. However, the CAD schemes proposed in all the previous studies [5], [8], [10] did not thoroughly examine the relationship between nodule size and malignancy in their proposed methodologies/approaches to solve the nodule classification problem.

To perform our preliminary analysis and throughout this study, we used 1,018 CT scans from the public Lung Image Database Consortium and Image Database Resource Initiative (LIDC-LDRI) dataset, which were collated and released by the National Institutes of Health (NIH) [11]. The nodule diameters in the LIDC-IDRI dataset range between 3 to 30 millimeters (mm). Figure 1 displays the nodule diameter distribution of all the nodules in the LIDC-LDRI dataset. From Figure 1, we observe that the malignant nodules in the dataset have diameters that typically exceed 12 mm. On the other hand, the benign nodules have diameters that are generally less than 5 mm. These two distributions describe the nodules that are "easy" to classify based on their nodule size/diameter. Namely, by applying a simple thresholding operator (i.e., nodule is malignant if size > 12 mm, benign if size < 5 mm), one can likely obtain a good/reasonable nodule classification score. However, the nodule sizes between 5 to 12 mm including the intersection of the two malignant and benign distributions in Figure 1 (at 8 mm) represent the nodules that are "difficult" to classify in this dataset. There is no straightforward/easy way to classify these nodules, and a novel methodology is required to more accurately classify the nodules in this size range, which has not been thoroughly examined in the previous studies.

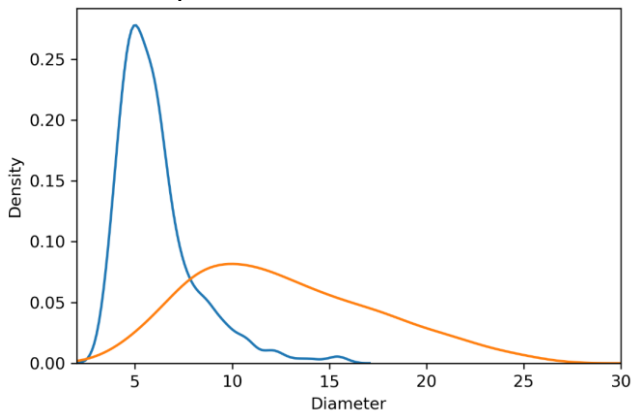

**Figure 1.** The nodule diameter distributions of all malignant and benign nodules within the LIDC-IDRI dataset.

Problems of this nature are generally known as multi-scale problems in the field of computer vision. Over the last few years, researchers have proposed various scale-invariant models to resolve this issue [12]–[14]. Most of these methods transform an image into multiple sizes, and these images are then forwarded to a classification network. The transformation step can be applied at the beginning/start, middle stage, or at the end of the network. More recently, Shen et al. [15] applied a Multi Crop Convolutional Neural Network (MC-CNN) strategy to capture multi-scale features for the lung nodule classification task. Using the MC-CNN method, max-pooling, cropping, or both max-pooling and cropping steps are applied to the input features of the networks. The output of these operations are feature maps with different spatial sizes. However, the issue with this and other scale-invariant based methods is that the max-pooling operation throws away many features, and only retains/selects the maximum result. Consequently, the resulting output resolution of the features is considerably reduced.

In this paper, we present a novel and completely different strategy/approach to address the nodule size and classification problem. In our new approach, in contrast to the previous methods, instead of reducing the resolution of the features, we increased the local receptive fields of the convolutional filter to cover a wider image area without increasing the number of parameters. Our new approach is based on the Dilated Convolution or Atrous Convolution Neural Network [16], [17]. The Dilated Convolutional Neural Network has been applied successfully as a replacement for pooling in semantic image segmentation [18], generic image classification [19], sound wave synthesis [20], and machine translation [21]. To the best of our knowledge, this is the first time that this network is combined with a gating mechanism on a medical imaging based classification problem. Similar to [15], we use multiple filters to capture multiple-scale features. However, in contrast to [15], we do not reduce the feature resolution and/or discard any features.

Moreover, we also designed a novel context-aware sub-layer to guide the features through the network layers, as depicted in Figure 2. The context-aware sub-layer generates signals that are responsible for closing or opening the gate that is located in front of each dilation neural network. This gives the network the ability to choose the right dilation for each nodule, depending on the nodule size as depicted in Figure 2. Thus, our Gated-Dilated (GD) network is based on the principle of using multiple dilated convolutional neural networks to capture the scale-invariant features, as well as a new gating mechanism to guide these features in the network. We will describe the methodology and inner workings/mechanisms of our new GD network in detail in the Methods section (i.e., Section III). Overall, our main contributions in this study can be summarized as follows:

1) We propose a new multiple dilated CNN to capture scale-invariant features that more accurately classify the lung nodules as benign or malignant.
2) We propose a novel context-aware sub-layer to guide the features between the dilated convolutions.
3) We performed a comprehensive validation of the GD network on the public and comprehensive LIDC-IDRI lung dataset of 1,018 CT scans and achieved state-of-the-art results.

The rest of the paper is organized as follows: The Related Work section that describes the background of the paper is presented in Section II. The Methods section is presented in Section III, which describes the proposed GD network in detail and also describes the experimental setup and methodology. The Results are described in Section IV.

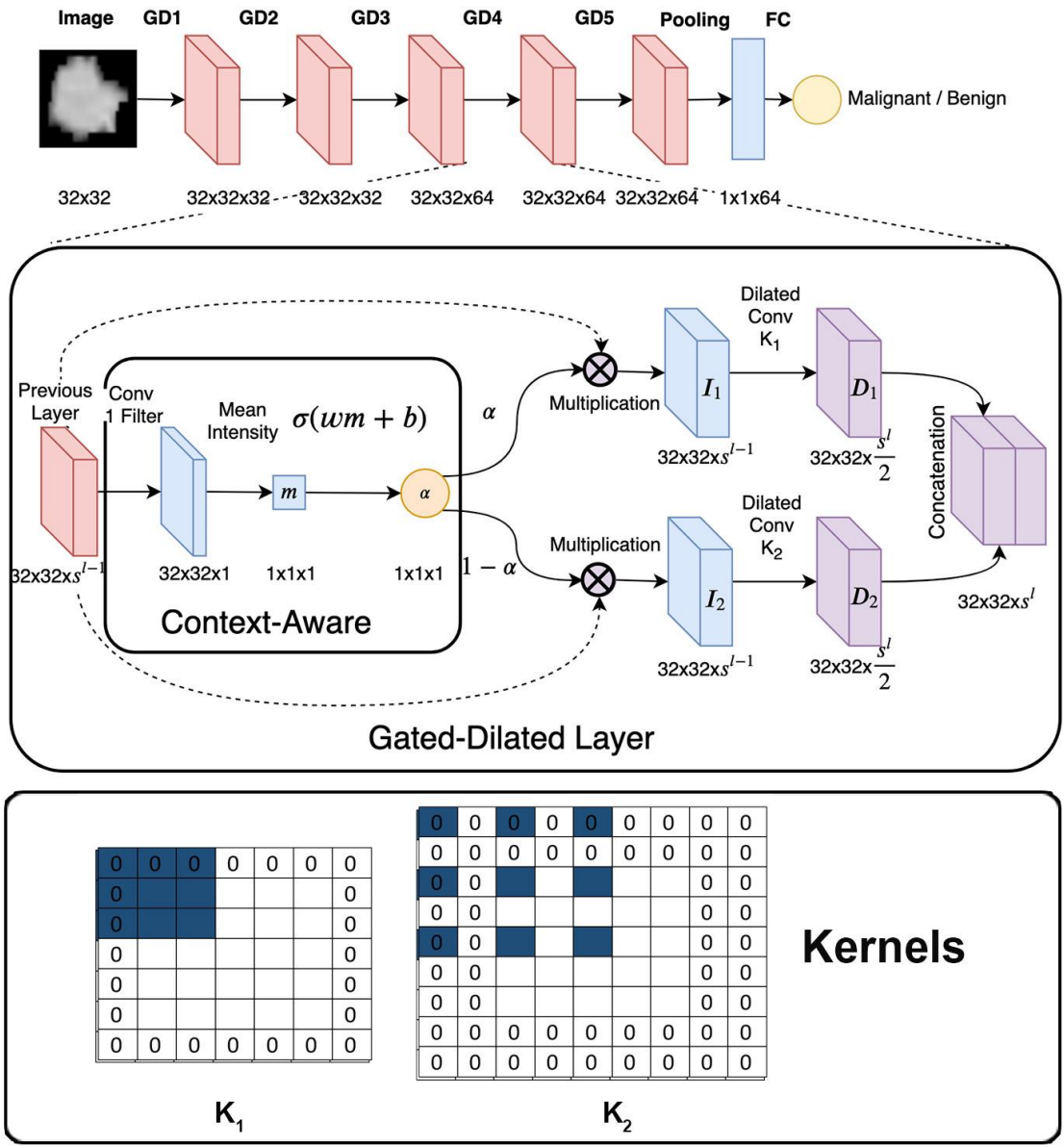

**Figure 2.** Illustration of our new Gated-Dilated (GD) network framework. The GD network consists of five GD layers, a Pooling layer, and a Fully-Connected layer. The design of each GD layer is shown in the magnified image (below). All hyperparameters in the GD layer are fixed except the $S^l$ value which controls the number of output channels in the $l^{th}$ layer. We set the $S^l$ value for GD1, GD2, GD3, GD4, and GD5 to 16, 16, 32, 32, and 32 respectively. The bottom part of the diagram illustrates how the kernels $K_1$ and $K_2$ operate on 5x5 input features. The shaded area in blue depicts the convolution area/region. Two dilated convolutions, $K_1$ and $K_2$ are designed to extract image features of two scales for smaller and bigger nodules, respectively as shown by their receptive fields in this figure. Note that the kernels first pad the input with zeros to maintain the same spatial size after the convolution.

Finally, the Discussion and the Conclusions of the study are provided in Sections V and VI, respectively.

## II. RELATED WORK

One of the few articles that analyzed other types of deep learning networks instead of CNN is Sun et al. [22]. In ref. [22], the authors compared three different deep learning methods, namely CNN, Deep Belief Networks (DBNs), and Stacked Denoising Auto-Encoder (SDAE). The authors extracted 134,668 samples from the LIDC dataset. They obtained so many samples by enlarging their dataset through data augmentation procedures, namely rotation of the extracted regions of interest (ROIs) to four different directions. Their results showed that CNN outperformed DBN and SDAE, in terms of the accuracy and area under the receiver operating characteristic (ROC) curve (AUC) measurements.

Deep residual neural networks (e.g., Resnet) have been analyzed on generic datasets (e.g., CIFAR-10, CIFAR-100 and ImageNet) to tackle the vanishing-gradient problem in recent studies in the literature [7]. By introducing "skip connections" in the network architecture, this enabled the deep neural network to have up to 1,000 layers. Nibali et al. [8] applied three Resnet-18 network topologies/structures to the nodule classification task, whereby each of the three networks were applied to different views/planes of the nodule, namely the axial, coronal, and sagittal views. The outputs of the three different networks were then passed to a fully-connected multi-layer perceptron network. Before training their network with the LIDC-IDRI dataset, they first trained it with the generic and non-medical CIFAR-10 dataset, to analyze the effect of transfer learning on the accuracy of the malignancy classification.

Hussein et al. [23] proposed using a three-dimensional (3D) CNN for lung nodule risk stratification, to utilize volumetric information in CT scans. Similar to ref. [8], the authors examined a transfer learning strategy to train their CNN. However, unlike ref. [8] their CNN could not be trained on 2D images, such as CIFAR-10 or Imagenet due to its 3D topology/structure. Thus, the authors trained their 3D CNN on a sports dataset comprising of 1 million videos [24].

Inspired by evolutionary intelligence based methods, Silva et al [25] examined an evolutionary CNN to classify lung nodules. Namely, the authors combined Particle Swarm Optimization (PSO) [26], genetic algorithms [27] and CNNs to solve the nodule classification problem. First, the authors combined PSO and Otsu's algorithm [28] to segment the nodules. The authors subsequently used a CNN to classify the nodules as malignant or benign. Designing a CNN requires knowledge of hyperparameters, such as the number of kernel filters and the number of neurons. Usually, these hyperparameters are fine-tuned by hand or by applying automatic algorithms to select the best hyperparameters, e.g., Grid-Search or Random Sampling [29]. In contrast to the previous methods, the authors examined genetic algorithms to select the best hyperparameters of a CNN.

In ref. [30], Liu et al. proposed a multi-view CNN applied to a lung nodule detection problem. The authors utilized three scales and four views, resulting in 12 different images altogether. Each image was used for training a CNN separately, following which, the authors combined all the models and re-trained the CNN again. During the preprocessing stages, the authors applied linear interpolation, normalized spherical sampling using an icosahedron positioned at the nodule centers, and nodule radius approximation by thresholding. Using their proposed approach, the authors achieved a high classification rate of 92.1%.

## III . METHODS

In this section, we describe our new Dilated Convolution and Gated-Dilated sub-network method in detail. In Section III.A, we first formulate and explain the dilated convolution network and our new Gated-Dilated Layer. We then explain the overall network architecture in Section III.B.

### A. THE GATED-DILATED LAYER

The dilated convolution operation is a variation of the standard/regular convolution operation [18]. Namely, if we have input features, $X$ (e.g., image pixels) and a filter, $K$ we can write the convolution operation as follows:

$$(X * K_r)(i,j) = \sum_m \sum_n X(i - m * r, j - n * r) K(m, n) \quad (1)$$

where $r$ is the dilation rate. The advantage of using dilated convolutions over regular convolutions is that the dilated convolution covers a wide range of input features. For example, if the filter size is 3x3 with a dilation rate of two, the receptive field covers an area of 5x5 without increasing the number of parameters and without reducing the resolution of the features. Thus, dilated convolutions can be used instead of max-pooling, but without reducing the output resolution or losing any information. In each GD layer, we used two dilated convolutions $K_1$ and $K_2$ with dilation rate 1 and 2, respectively and kernel size of 3x3. The $K_1$ filter works as a conventional convolution filter which is good for capturing small patterns, whereas the second filter $K_2$ covers a wider area of features and is therefore suitable for capturing bigger patterns. Both kernels are applied with stride = 1, and zeros are padded to the input before applying the convolution to maintain the same spatial size as the input.

Another unique and novel contribution of our approach is that we present a novel Context-Aware sub-network to guide the features between the multiple dilated convolutions. Unlike other methods in the literature and previous studies, our Gated-Dilated (GD) network has a Context-Aware sub-network that analyzes the context of other input features and generates attention signals to guide the features between the multiple dilated convolutions. The architecture of the Context-Aware sub-network is depicted in the magnified image in Figure 2.

The context-aware sub-network consists of a single 3x3 convolution filter with a Rectified Linear Unit (ReLU) activation function, followed by a global average pooling layer. The function of the pooling layer is just to reduce the number of parameters and features before we feed them to the next stage, and does not reduce the resolution of the input features, $X$. The kernel size of the global average pooling equals the spatial size of the input feature which is 32. Thus, the output of the global pooling is just a scalar value. Finally, the scalar value is passed through a single neuron with a sigmoid activation function. The output of the sigmoid activation function, $\alpha$ represents the attention signal, which will be used to guide the features between the convolutions. As $\alpha$ is conditioned on input, $X$, we can express the function/operation of the Context-Aware sub-network as:

$$P(\alpha|X) \quad (2)$$

We use the gate function to control the flow of features through the dilation filter. The attention gate function is simply an element-wise multiplication between input *X* and the scalar *α* signal. The gate function uses a soft gate instead of a hard gate; the advantage of using a soft gate is that *α* is continuous (i.e., not a binary/hard threshold), and can be any value between 0 and 1. In our network architecture (see magnified image of Figure 2), we have two attention gates corresponding to two different dilation rates, applicable to smaller and bigger nodules, respectively. Thus, if the value of *α* is close to 1, the attention (or weight) will be on the convolution with the first dilation rate, $I_1$ rather than the second dilation rate $I_2$, as follows:

$$I_1 = \alpha X \qquad (3)$$

$$I_2 = (1 - \alpha)X \qquad (4)$$

The outputs of the attention gates, $I_1$ and $I_2$ are then passed to two dilated convolution filters, $K_1$ and $K_2$, respectively. In our method, we implemented these two filters to have a 3x3 kernel size and a ReLU activation function. Thus, the dilated convolutions are performed as:

$$D(I_1, K_1)_1 = (I_1 * K_1)(i, j) \qquad (5)$$

$$D(I_2, K_2)_2 = (I_2 * K_2)(i, j) \qquad (6)$$

The output of these two filters, $D_1$ and $D_2$ are used to capture the variation in the nodule sizes, whereby $D_1$ acts as a regular convolution that is suitable for smaller nodules, whereas $D_2$ is suitable for bigger nodules. Finally, the results of $D_1$ and $D_2$ are channel-wise concatenated and passed to the next layer.

### B. OVERALL NETWORK ARCHITECTURE

Figure 2 depicts the entire network architecture or top-level diagram of our new GD network. We applied five consecutive GD layers with 32, 32, 64, 64, and 64 channels, respectively. For each GD layer, the number of channels is equally divided/separated between the two dilated convolutional layers. For example, in the first GD layer, the number of channels is 32. This means that the convolutional layer with dilation rate of 2 has 16 channels, and the convolutional layer with dilation rate of 1 also has 16 channels. Therefore, after concatenating the outputs of the two layers, the total number of channels sums up to 32. It is common in the deep convolutional neural network to increase the channels as you go deeper [7], as the high-level features are more detailed than the generic low-level features; thus, the deeper layers require more filters.

To prevent the network from "overfitting" on the training dataset, multiple dropouts [31] were implemented in our network architecture. Networks that are less likely to overfit the training dataset are capable of better generalization on the independent (i.e., unseen) testing dataset. Thus, in our implementation/framework, we placed the first and second dropout layers after the second and fourth GD layer, respectively with a dropout rate (or probability) of 0.25. Another dropout layer was placed after the fifth GD layer with a dropout rate of 0.5. We increased the dropout rate (from 0.25 to 0.5) as we proceed deeper into the network as deeper layers have more parameters than shallower layers, which makes them more prone to overfit on the training dataset.

After the five GD layers, we used global max-pooling to summarize the feature space. The global max-pooling operation computes the maximum value of all features in the feature maps. That is, for each channel, we obtain a single value which is the maximum value of the channel. For example, the last GD layer has 64 channels; thus, after applying global max-pooling, we obtain 64 features altogether. We used max-pooling to reduce the number of trainable parameters in the fully-connected network layer, which comes after the max-pooling layer. This is helpful as the size of the dataset is small (see Sections III.C and III.G), which may lead to overfitting if the number of parameters is huge. These features are then passed/transmitted to the last layer, which consists of a single neuron with a sigmoid activation function. The output of the sigmoid activation function is the probability of the nodule in question being malignant. Namely, if the output of the sigmoid exceeds 0.5, this means that the network predicts that the nodule in question has a higher probability of being malignant. Otherwise, it is more likely to be benign. As we have only one output and a two-class (malignant/benign) classification problem, we used the binary cross-entropy loss function. We also optimized the network parameters using the Adam optimizer [32].

### C. DATASET DESCRIPTION

The dataset used in this work is the LIDC-IDRI dataset [11] released by the National Cancer Institute, NIH. The LIDC-IDRI dataset is the largest and most comprehensive public lung nodule dataset. It consists of 1,018 CT scans collated from 1,010 patients altogether. The large size of the dataset and its public availability makes the LIDC-IDRI dataset suitable for developing and comparing/validating different deep learning based methods and it is a dataset that is frequently studied in the literature [3], [15], [22]. As the images were collated from four different institutions using different CT scanners, there is a wide variation of image parameters within this dataset. For example, the image resolution, namely the pixel spacing in mm of different CT scanners are different; the slice thickness of the CT scans also range from 0.45 to 5.0 mm. Using a diverse dataset like LIDC-IDRI to develop algorithms has the advantage that the algorithms developed can be robust to unseen/generalized data as they have been trained on a diverse dataset.

The malignancy suspiciousness of each nodule in the LIDC-IDRI dataset was rated by four experienced radiologists. First, the radiologists annotated all nodules in each CT scan, whereby the nodule boundaries were provided in individual XML files. Only nodules with diameters between 3 and 30 mm were annotated. Similar to previous studies, we observed that there are wide variations/variabilities in the nodule annotations [3], [33]. Namely, although the scans were annotated by four experienced radiologists altogether, only some of the nodules in the dataset were annotated by the majority of the radiologists (i.e., by at least three out of four of the radiologists). Thus, similar to our previous work in ref [3], we used a gold standard of the majority or at least three out of four radiologists to define a nodule. To group the annotations that have the same nodules, we used a nodule size report [34] similar to ref. [8].

The radiologists also rated the malignancy suspiciousness of the nodules from 1 to 5, indicating an increasing degree of malignancy suspiciousness (namely, 1 represents a benign nodule; 5 is highly malignant). We combined the radiologists' ratings by taking the median of the malignancy levels: ratings less than three were considered as benign, whereas ratings above 3 were considered as malignant. Similar to previous similar studies [15], [22], we also excluded nodules that had ratings of exactly 3 as these nodules have an indeterminate malignancy status. We also excluded nodules with ambiguous IDs from our dataset. In this way, we obtained 848 nodules altogether of which 442 are benign and 406 are malignant.

### D. PREPROCESSING
First, to segment the nodules from their surrounding regions, we used the nodule annotations of the four radiologists. Second, to avoid partial volume effects caused by the different CT scanning protocols across different vendors, we used trilinear interpolation to normalize the CT scan volumes, resulting in isotropic resolution in all three (*x*, *y*, and *z*) dimensions. We then extracted a 32 by 32 millimeter square region about the center of each nodule, which we provided to the input of the GD network (see input image of Figure 2).

### E. DATA AUGMENTATION
Deep learning based methods are generally data hungry and require training with large datasets to generalize well on unseen testing datasets. Thus, we trained our GD network with augmented datasets to improve its generalization capabilities given the limited number of training samples. From each nodule, we extracted three 2D views from the axial, coronal, and sagittal planes. We applied four rotation angles (i.e., 0º, 90º, 180º, and 270º) to each 2D view. We also applied Gaussian blurring with scale, $\sigma = 1$ on the axial, coronal, and sagittal views of the 0º image, and a recent article showed that training the CNN on Gaussian-blurred images improved the overall diagnosis of thorax diseases in 2D chest x-rays [35]. Thus, we obtained 15 different images altogether of each nodule through the data augmentation procedure. We used all 15 images to train the networks. During testing, we obtained the output of the three views separately and averaged their results to obtain the final output.

We also normalized all the images in our dataset to have zero mean and unit variance using the standard score (or *z*-score). The mean and the standard deviation of the training dataset was computed, and normalization with the same mean and standard deviation was applied to the images in the testing dataset.

### F. BASELINE METHODS/COMPARISONS
Besides the GD network, we designed and implemented three other networks to form the baseline methods or comparisons for our experiments and compared them with our method's performance. The first baseline method is a conventional CNN model that has the same number of layers as our model and the same number of channels in each layer. That is, we implemented a network architecture similar to Figure 2; however, instead of using the GD layer in the magnified image in Figure 2, we used a conventional CNN instead. The purpose/objective of using this method as a baseline comparison is so that we can analyze the mechanism of the GD layer and examine whether the GD layer can successfully differentiate between malignant/benign nodules (using attention gates to guide the dilated convolutions), compared to the conventional CNN constructed with the same parameters.

Besides conventional CNN, we designed another two ablation studies to analyze the contributions of the dilated convolution and the Context-Aware Sub-network, respectively. The first study is called GD-No-Dilation, which is similar to our proposed method, except both $K_1$ and $K_2$ have the same dilation rate of one. This study will help us understand whether having different dilation rates (of 1 and 2 in the original network architecture) will have any benefits/effects on the obtained results or not. In the second ablation experiment, we designed a model called GD-No-Gate, which is similar to our proposed method except that there is no gating or Context-Aware Sub-network in the network architecture. This ablation study will shed light on whether the Context-Aware Sub-network is useful in guiding the features through the network layers.

Additionally, we implemented a state-of-the-art lung nodule classification model called Multi-Crop Convolutional Neural Network (Multi-Crop) [15]. We also implemented other state-of-the-art generic image classifiers models, namely Resnet-50 [7] and Densenet-161 [36] pre-trained with the Imagenet dataset [37] to improve their performance by transfer learning. As Resnet and Densenet work with "Red Green Blue" or RGB images, we duplicated the nodule grayscale image to all three RGB channels. We also resized the nodule images to 224x224 to meet the input requirement dimensions of the Resnet and Densenet networks. Regarding the network topology of Resnet and Densenet, we only modified the last layer by replacing it with a single neuron to meet the requirement of the two-class (malignant/benign) nodule classification problem. We tested two methodologies of transfer learning: In the first one, we fine-tuned all the parameters of the network using the LIDC-IDRI dataset, and we refer to these networks throughout the paper as Resnet-Full and Densenet-Full, respectively. In the second methodology, we only fine-tuned the last layer, whereas the other layers were trained using the Imagenet dataset, and we refer to these networks throughout the paper as Resnet and Densenet. By fixing all the layers except the last one during training, we reduce the risk of overfitting the data. Moreover, the gradients only pass through the last layer, hence the gradients will not explode or vanish.

For all baseline methods except Multi-Crop, we used the same loss function and optimizer as the GD network and duplicated all other hyperparameters of the GD network (also see Section III.G on Experimental Design and Evaluation). This was performed to maintain all the parameter settings across the nine methods as much as possible, to ensure a fair and comparable comparison of our GD network with these state-of-the-art baseline methods. For Multi-Crop, we applied the same settings as ref. [15] including maintaining the input size at 64x64x64 as described in ref. [15].

### G. EXPERIMENTAL DESIGN AND EVALUATION

Our GD network implementation was based on Pytorch [38]. We ran our experiments using a NVIDIA Titan X Pascal GPU. The GD network was trained with a batch size of 256 for 50 epochs. We initially set the learning rate to $1.0\times10^{-3}$ and decreased it to $1.0\times10^{-4}$ after the 20th epoch, similar to what was performed for training Resnet in refs. [7]. We set the learning rate to be higher initially so that the gradient descent is faster at the early stages of training; then, we reduced the learning rate so that the gradient descent algorithm will not overshoot and skip the global minimum error. We set the hyperparameters of the Adam optimizer [32], $\beta_1$ and $\beta_2$ to the default values of 0.9 and 0.999, respectively. We initialized all the weights using the Xavier uniform initialization [33] except the network bias values, which were initialized to zero. We applied these settings to the GD network as well as to the other baseline methods to ensure that fair performance comparisons could be made across the different methods except Multi-Crop, whereby we set the hyperparameters as described in ref. [15].

We validated all nine methods using a ten-fold cross-validation method whereby the sum of the 406 malignant and 442 benign nodules where randomly divided into 10 exclusive partitions (or subgroups). In the random division, we maintained the ratio of benign to malignant nodules as much as possible across all 10 folds to ensure that the distribution of labels in each fold was approximately equal. Thus, after the division into 10 folds, the ratio of benign examples in each fold was around 51%. In each validation (training and testing) cycle, nine subgroups were used to train the network, and the trained network was then applied to the remaining subgroup. For each testing sample/nodule, the network generated an output score ranging from 0 to 1. A higher score indicates a higher probability/likelihood of the nodule being malignant. This process was iteratively executed 10 times using the 10 different combinations of subgroups. In this way, each of the 848 nodules was tested once with a corresponding network-generated probability score.

### IV. RESULTS

Figure 3 displays four ROC curves of the GD network, Resnet, Densenet, and Multi-Crop networks that were generated by a well-established software package for computing ROC curves and corresponding AUC values, namely ROCKIT[TM] [39]. The ROC curves for the other five baseline methods were omitted to enhance the readability/ visibility of the four curves in Figure 3; however, the corresponding AUC, accuracy, precision, and sensitivity results of all methods are summarized and tabulated in Table 1. We observe from Figure 3 that Resnet, Densenet, and Multi-Crop have almost identical ROC curves and very similar true positive rates across all false positive rates with Densenet slightly outperforming Resnet and Multi-Crop. The results also show the superiority of the GD network, which significantly outperforms all other methods across all false positive rates.

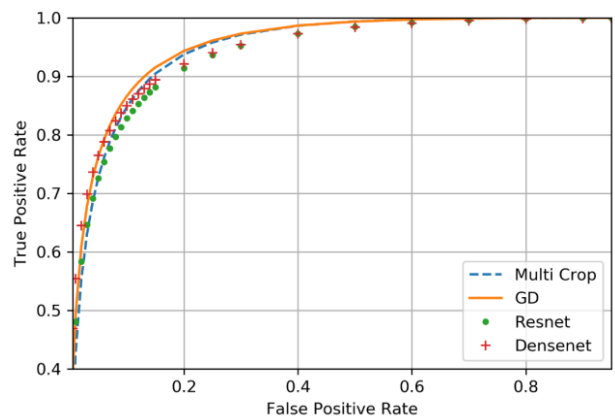

**Figure 3.** Comparisons of four receiver operating characteristic (ROC) curves of our GD network, Resnet, Densenet and a Multi-Crop network with very similar hyperparameters and network architecture as our GD network. It can be observed that the ROC curve of our GD network is very competitive compared to the other baseline methods.

The AUC, accuracy, precision, and sensitivity results of all nine methods are tabulated and compared in Table 1. The results in Table 1 show that the GD network outperforms the other eight baseline methods by significant margins in terms of accuracy = 92.57%, AUC = 0.9514, and precision = 91.85%. Although the CNN network has a marginally higher sensitivity result than the GD network, the difference in performance is very small (i.e., 92.67%-92.21%=0.46%). Densenet outperforms CNN and Resnet across all performance metrics except sensitivity and precision, whereby it is outperformed by CNN and Resnet, respectively. If we compare the results of CNN with Resnet, we observe that Resnet outperforms CNN across all performance metrics except sensitivity. Additionally, we classified the nodules without the segmentation mask and the results changed marginally (i.e., <1%).

**Table 1.** Classification performance comparisons of the proposed GD network with the eight baseline comparison methods. We also adopted the reported results of DenseBTNET, GBRT, and HSCN from the published papers.

| Methods | AUC | Accuracy | Precision | Sensitivity |
|---|---|---|---|---|
| DenseBTNET[40] | 0.9315 | 88.31% | --- | --- |
| GBRT[10] | 0.91 | 84% | 0.874 | 0.886 |
| HSCNN[41] | 0.856 | 84.2% | 0.889 | 0.705 |
| Multi-Crop[15] | 0.9480 | 89.27% | 0.8637 | 0.9212 |
| Resnet | 0.9410 | 90.45% | 0.8889 | 0.9026 |
| Resnet-Full | 0.9198 | 87.62% | 0.8525 | 0.8966 |
| Densenet | 0.9481 | 90.90% | 0.8887 | 0.9031 |
| Densenet-Full | 0.9332 | 87.50% | 0.8472 | 0.9015 |
| **GD** | **0.9514** | **92.57%** | **0.9185** | 0.9221 |
| CNN | 0.9397 | 90.33% | 0.8707 | **0.9267** |
| GD-No-Dilation | 0.9485 | 90.21% | 0.8854 | 0.9138 |
| GD-No-Gate | 0.9353 | 90.68% | 0.8978 | 0.9089 |

The ablation studies in Table 1 (i.e., GD-No-Dilation and GD-No-Gate) show the importance of applying dilation and gating on the overall results: GD outperforms both GD-No-Dilation and GD-No-Gate across all performance metrics. The results also demonstrate the superiority of Resnet and Densenet over Resnet-Full and Densenet-Full, respectively. Resnet and Densenet outperform Resnet-Full and Densenet-Full across all performance metrics, which means that fine tuning the last layer and keeping all other layers fixed is better than tuning all the layers. This is especially true in this study, whereby the dataset size is comparatively smaller to the size of the Resnet and Densenet models.

To analyze the performance of our GD network on nodules of different diameters within the LIDC-IDRI dataset, we compared the accuracies of all models on different nodule diameters. Similar to Figure 3, to enhance the readability/visibility of the results, we only plotted the accuracies of the GD network, Resnet, Densenet, and Multi-Crop in Figure 4. As expected, the nodules with large diameters of 13 to 25 mm are easily classified by all four models except Multi-Crop. This is because malignant nodules generally have bigger diameters/larger sizes than benign nodules as shown in Figure 1, and also confirmed by a recent study conducted on a very big dataset [5]. Similarly, benign nodules generally have smaller sizes than malignant nodules; thus, the very small nodules of 3 to 4 mm in diameter were accurately classified by all four methods in Figure 4 except Multi-Crop. Figure 4 also shows that the nodules that are difficult to classify are those that have diameters between 5 to 12 mm. The results show that our GD model outperforms all other methods on the difficult nodules within this diameter range. The accuracy of the GD network also exceeds or performs at least as well as all other methods across the range of small, medium and large-sized nodules.

### A. EXPLORATORY ANALYSIS OF THE ATTENTION SIGNAL

The attention signal $\alpha$ that we introduced in equation (2) and used in equations (3) and (4) has a critical effect on the overall output of the GD layer. As explained in Section III.A, $\alpha$ is used to guide the features between the different dilations based on the nodule size. Due to the significant/important role of $\alpha$, we conducted an experiment to study the relationship between the nodule area size and $\alpha$, and to also analyze whether $\alpha$ effectively guides the dilated convolution filters between smaller and bigger nodules, respectively. Thus, after training the GD model, we forwarded the 2D nodule images through the network and recorded the attention signal $\alpha$ for each image. We also estimated the nodule area by multiplying the width and the height of the nodule in each 2D image and examined the relationship between $\alpha$ and the estimated nodule area.

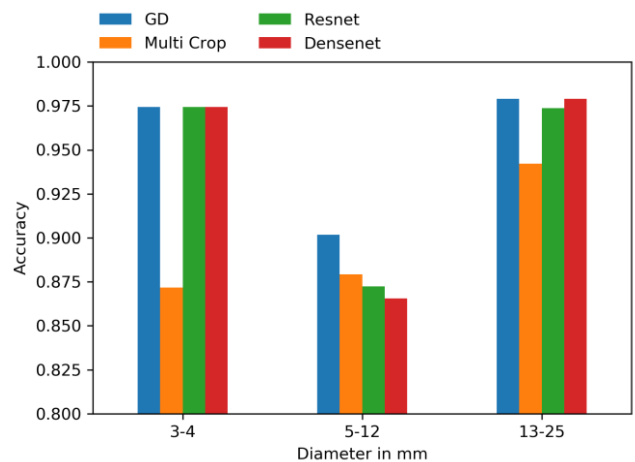

**Figure 4.** Accuracies of the GD network, Resnet, Densenet, and Multi-Crop across small (3-4 mm), medium (5-12 mm) and big (13-25 mm) nodule diameters within the LIDC-IDRI dataset.

The results of our analysis performed on all five GD layers in our network architecture are depicted in Figure 5 (layers 1 to 5 depict increasingly deeper GD layers as we progress through the GD network architecture in Figure 2). We can observe several interesting and unique findings of the results in Figure 5. First, we observe that the relationship between $\alpha$ and the nodule area is approximately linear for all layers except layer 4. The linearity between $\alpha$ and the nodule area follows/agrees with the linear relationship assumption that we hypothesized in equations (3) and (4) of Section III.A. Another interesting observation is that the value of $\alpha$ never becomes close to zero or one; that is, for most nodules, $\alpha$ is between 0.4 and 0.6. This result means that both dilation rates 1 and 2 of our proposed model are useful to extract features from the nodules, which also explains and confirms the good classification results obtained for our GD model in Figure 4. We also observe that as the nodule area increases, $\alpha$ decreases, which means that the dilation rate of 2 gets more attention as the nodule size increases. This trend is observed for all the layers except layer 4. In layer 4, which is the second-last GD layer, $\alpha$ for all nodule sizes is always less than 0.5 (specifically, 0.46 or less), which means that the model prefers to give a higher attention to dilation rate 2 compared with dilation rate 1 as we go deeper in the network. As we progress to deeper layers, global features (extracted by dilation rate 2) are emphasized more than local features, which explains the higher attention signal given to dilation rate 2 in layers 4 and 5.

Table 2. Pearson's correlation coefficient between the attention signal, $\alpha$ and the nodule area for all five GD layers in Figure 2.

| Alpha | Conv1 | Conv2 | Conv3 | Conv4 | Conv5 |
|---|---|---|---|---|---|
| Correlation | -0.96 | -0.93 | -0.97 | 0.57 | -0.93 |

We also measured Pearson's correlation coefficient between the attention signal, $\alpha$ and the nodule area. Table 2 tabulates Pearson's correlation coefficient between the two variables for all five GD layers. Except for layer 4, a strong negative linear correlation of -0.93 or less exists between the two variables. For layer 4, Pearson's correlation coefficient was around 0.58, which demonstrates only a moderate linear correlation between the two variables. These results confirm our previous finding that there is a linear relationship between the nodule area and $\alpha$ for all GD layers except layer 4.

## V. DISCUSSION

In this paper, we proposed a new GD Network to classify lung nodules as either malignant or benign. The intuition behind this novel architecture is to work around the high variation in the nodule diameter, which ranges from 3 to 30 mm. Extremely big or small nodules are easy to classify as very big nodules are usually malignant, whereas smaller ones are usually benign. As shown in Figure 1, the nodules with diameters between 5 and 12 mm are harder to classify than the extremely big or small nodules. Thus, we proposed using multiple convolutional filters in parallel, whereby one filter captures the local features and the other the global features. Instead of using multiple pooling layers to obtain different scales of the feature representations, we used dilation convolution with different dilation rates to capture the wide range of nodule diameters. By increasing the dilation rate, we can cover a wide area of the features without the need to scale down the input features or reduce the feature resolution. The output from the two dilation convolutions combined using a concatenation function are passed to the next layer. In this way, different feature resolutions corresponding to different dilation rates are captured across the low and high level features as the network progresses to deeper layers.

Another novel and unique aspect of our network architecture is that we designed a Context-Aware sub-network to guide the features between the different dilation convolution filters. The Context-Aware sub-network is conditioned on the input features, which means that it first analyzes the input features. Based on the analysis, the Context-Aware sub-network generates an attention signal that controls the attention gates that are located in front of each dilated convolution filter as shown in Figure 2. The intuition is to let the network decide which path the features should take and how much each dilation rate as proportioned by the attention signal, should contribute to the output. A linear relationship was observed between the nodule size and attention signal generated by our network architecture, which indicates that the Context-Aware sub-network works effectively to guide the features to the right attention gate/dilation rate based on the nodule size. This is further confirmed by the results in Figure 4 that show that our GD network performs considerably better than the other baseline methods to classify nodules in the difficult size range (of 5-12 mm). Although we applied our GD network for lung nodule classification, the same principle could be applied to different problems where multi-scale objects exist.

Another contribution of this paper is that we used eight baseline comparison methods including Multi-Crop, Resnet, and Densenet, which are approaches that produced state-of-the-art results in the literature for the nodule classification task. Our results demonstrate that our new GD network significantly outperformed Multi-Crop, Resnet, Densenet, and the conventional CNN for almost all performance metrics including AUC (0.9514±0.0078), accuracy (92.57%±2.47) and precision (0.9185±0.0454). We maintained all hyperparameter settings as much as possible between the nine examined methods to ensure that valid performance comparisons and conclusions could be derived from the obtained results.

The results of this study are very timely, especially in this era of big data and lung radiomics [42]. The previous conventional lung nodule classification and detection methods used hand-crafted features, such as shape and texture based features to classify and detect nodules [43]. However, the process of designing/selecting relevant features is difficult and time consuming and rely on researchers' prior knowledge/expertise. Deep learning algorithms have the ability to address the limitations of current CAD schemes. First, deep learning algorithms can automatically extract meaningful/relevant features, thus eliminating the requirement of prior knowledge to derive useful hand-crafted features including heuristic descriptors. In our proposed method, our new GD network was able to extract relevant multiscale features without reducing the overall feature resolution. Second, in this era of big data, large amounts of data are readily available, which could extend to tomographic raw data (namely, "rawdiomics") [42]. The ability of deep learning based methods to handle large-scale datasets and automatically generate features is a requirement in this ever-changing era.

The current CAD schemes for lung nodule detection and classification including R2 Technology's commercialized ImageChecker CT Lung CAD scheme also produce high false positive rates/many false detections that distract radiologists. Furthermore, the performance of current CAD schemes is very uncertain and varies from one dataset to another, thus making the validity/reproducibility of CAD schemes a highly controversial topic [3], [44]–[46]. Subjective evaluation of CT images also lag in specificity especially when it comes to differentiating benign from malignant lung etiologies [42]. Thus, new approaches incorporating deep learning methods and image-based radiomics are required to improve the performance of current CAD schemes for lung nodule classification so that they can be integrated into clinical practice. The results of our novel GD network are very encouraging. Namely, unlike the current CAD schemes that have high detection sensitivities, but have high false positive rates [3], [44]–[46], our method has very high AUC, accuracy, precision, and sensitivity results.

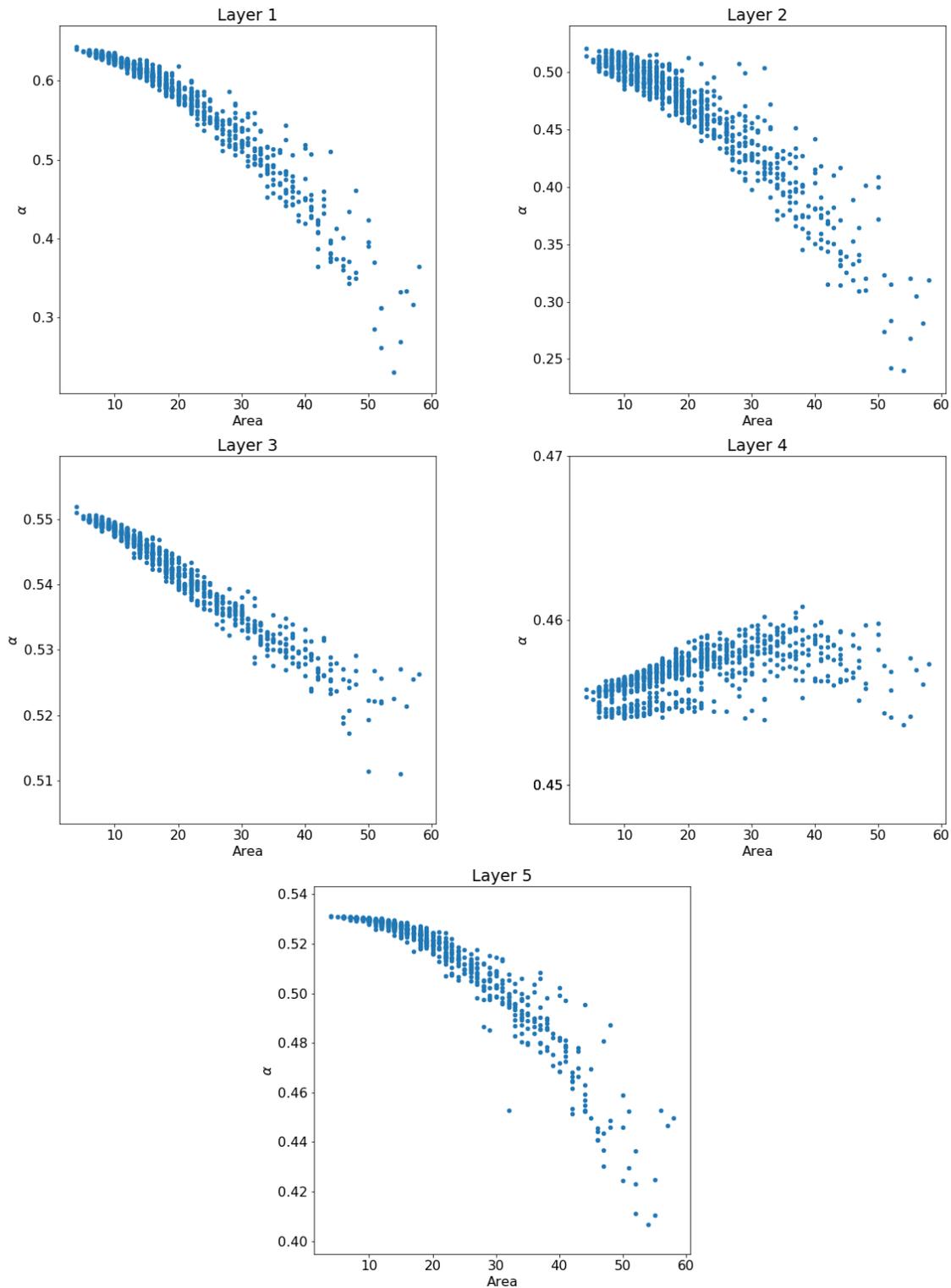

**Figure 5.** Analysis of the relationship between the area/size of the nodules and the attention signal, α across all GD layers of our proposed network architecture in Figure 2. Layers 1 to 5, respectively correspond to the increasingly deeper GD layers in Figure 2.

Although the results are very encouraging, we recognize that this is a preliminary study. First, our method is not a fully-automated method that can automatically extract/segment the nodule locations from whole CT scans and classify them as benign or malignant. Namely, our method currently requires an object detector model to identify the nodule locations before classifying them as benign/malignant. Although this is a current limitation of our method, other similar and recent studies for lung nodule classification also have this same limitation/drawback [15], [22]. In future, we plan to incorporate a fully-automated nodule classification scheme that can automatically detect the nodule candidates/locations similar to the CAD scheme for lung nodule detection that we proposed in our previous study [3]. Second, the size of the LIDC-IDRI dataset is a limitation for training deep learning algorithms as many hyperparameters in each layer require very large datasets to obtain good training results. However, we recognize that the LIDC-IDRI dataset is still the biggest available public lung CT dataset, which has been used in other similar studies for the nodule classification task [15], [22]. Thus, its usage is beneficial for valid and comparable performance comparisons with other similar methods in the literature.

## VI. CONCLUSIONS

In this study, we developed a new deep learning model based on GD networks for the challenging task of lung nodule benign/malignant classification. Our work focuses on the wide diameter variation of the nodules, which can range anywhere between 3 and 30 mm. To tackle this challenging problem, we proposed a GD Layer that has two dilated convolutions in each layer. Each of these convolutions has a different dilation rate to capture different nodule sizes. Moreover, the input features are guided between the two dilated convolutions by a new Context-Aware sub-network. This sub-network generates attention signals that guide the input features through the network architecture. The proposed model achieves state-of-the-art results on the LIDC-LDRI dataset and outperforms eight baseline state-of-the-art methods including Multi-Crop, Densenet and Resnet in terms of AUC, precision and accuracy. The results also demonstrate significant improvements in terms of classification accuracy on "difficult" medium-sized nodules, specifically the nodules with diameters between 5 and 12 mm. Analysis of the relationship between the attention signal and nodule area/size demonstrates that the Context-Aware sub-network works effectively to guide the features to the right attention gate/dilation rate based on the nodule size.

## ACKNOWLEDGMENTS
The authors gratefully acknowledge the support of NVIDIA Corporation with the donation of the Titan X Pascal GPU used for this research. The authors also acknowledge the National Cancer Institute and the Foundation for the National Institutes of Health, and their critical role in the creation of the free publicly available LIDC-IDRI database used in this study.

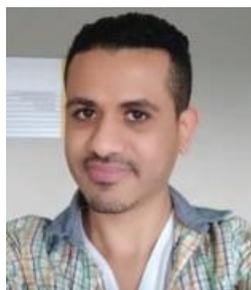

**Mundher Al-Shabi** is a PhD candidate at Monash University. He earned his Bachelor degree at Multimedia University, Malaysia in Artificial Intelligence. He is interested in applying deep learning in medical imaging problems.

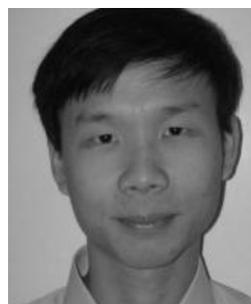

**Hwee Kuan Lee** received the Ph.D. degree in theoretical physics from Carnegie Mellon University, Pittsburgh, PA, in 2001. He is the head of the Imaging Informatics division in Bioinformatics Institute. His current research work involves developing and deployment of machine learning and deep learning algorithms for a wide range of applications, such as in computer vision, biomedical applications and in condensed matter physics.

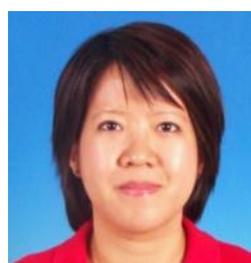

**Maxine Tan** is a Lecturer at the Electrical and Computer Systems Engineering Discipline, School of Engineering at Monash University. Her current research interests include medical image analysis, deep learning, and the development of Quantitative Imaging (QI) phenotype biomarkers for improving cancer screening, diagnosis and prognosis assessment.